\documentclass[journal]{IEEEtran}
\usepackage{amsmath,amsfonts}
\usepackage{array}
\usepackage{textcomp}
\usepackage{stfloats}
\usepackage{url}
\usepackage{verbatim}
\usepackage{graphicx}
\usepackage{arydshln}
\usepackage{cite}
\usepackage{orcidlink}
\usepackage{subfigure}
\usepackage{booktabs}
\usepackage{multirow}
\usepackage{bm}
\usepackage{epstopdf}
\usepackage{xcolor}
\usepackage{mathrsfs}
\usepackage[ruled]{algorithm2e}
\usepackage{amssymb}
\usepackage[figuresright]{rotating}
\usepackage{graphicx} 
\usepackage{subcaption}
\usepackage{color}
\usepackage{hyperref} 
\usepackage{authblk} 
\usepackage{hhline}
\newcommand{\thickhline}{\noalign{\hrule height 1pt}}

\begin{document}
\newenvironment{proof}{{\indent \it Proof:}}{\hfill $\blacksquare$\par}
\title{Learning from Expert Factors: Trajectory-level Reward Shaping for Formulaic Alpha Mining}


\author{Junjie Zhao$^{\orcidlink{0000-0003-0718-4183}}$,
Chengxi Zhang$^{\orcidlink{0009-0009-5876-5632}}$,
Chenkai Wang$^{\orcidlink{0009-0003-9194-2350}}$,
Peng Yang$^{\orcidlink{0000-0001-5333-6155}}$,~\IEEEmembership{Senior Member,~IEEE}
        
\thanks{This paper was produced by the IEEE Publication Technology Group. They are in Piscataway, NJ.}
\thanks{Manuscript received xxx, xxx. \textit{(Junjiezhao and Chengxi Zhang are co-first authors.) (Corresponding author: Peng Yang.)}

Junjie Zhao, Chenkai Wang and Peng Yang are with Southern University of Science and Technology, Shenzhen 518055, China (e-mail: zhaojj2024@mail.sustech.edu.cn; wangck2022@mail.sustech.edu.cn; yangp@sustech.edu.cn). Junjie Zhao is also with the University of Sydney (e-mail: jzha0632@uni.sydney.edu.au).

Chengxi Zhang is with Harvard University, Cambridge, MA 02138, USA (e-mail: chengxizhang@g.harvard.edu). 
}}



\maketitle

\begin{abstract}
Reinforcement learning (RL) has successfully automated the complex process of mining formulaic alpha factors, for creating interpretable and profitable investment strategies. However, existing methods are hampered by the sparse rewards given the underlying Markov Decision Process. This inefficiency limits the exploration of the vast symbolic search space and destabilizes the training process. To address this, Trajectory-level Reward Shaping (TLRS), a novel reward shaping method, is proposed. TLRS provides dense, intermediate rewards by measuring the subsequence-level similarity between partially generated expressions and a set of expert-designed formulas. Furthermore, a reward centering mechanism is introduced to reduce training variance. Extensive experiments on six major Chinese and U.S. stock indices show that TLRS significantly improves the predictive power of mined factors, boosting the Rank Information Coefficient by 9.29\% over existing potential-based shaping algorithms. Notably, TLRS achieves a major leap in computational efficiency by reducing its time complexity with respect to the feature dimension from linear to constant, which is a significant improvement over distance-based baselines.

\end{abstract}


\renewcommand{\arraystretch}{1.2}
\section{Introduction}
In quantitative finance, constructing a profitable investment strategy fundamentally relies on extracting informative signals from massive yet noisy historical market data, which is widely regarded as a signal processing problem\cite{shi2024saoftrl,zhao2019optimal,zhao2018mean}. Among these signals, \textit{alpha factors}\cite{kakushadze2016101}, which are quantitative patterns derived from these data, serve as structured indicators that guide asset return prediction, portfolio construction, and risk management.


The representations of alpha factor mining methods generally fall into two categories: parameterized learning models and mathematical formulations. The former can be either traditional tree models ~\cite{zhang2023openfe,zhu2022application,li2022research} or advanced neural networks models~\cite{wang2023accurate,liu2025multiscale,huang2020q,graves2012long,vaswani2017attention}.
Despite the capability of modeling nonlinear and high-dimensional dependencies, these models typically lack interpretability in terms of financial intuition and knowledge. This makes them difficult to validate and refine by human experts in real-world deployment, where risk controls are highly required.

In contrast, formulaic alpha factors are mathematical expressions that can be more easily understood by humans and thus offer strong interpretability. In early stage, alpha factors were often designed by financial professionals, thus could reflect well-established financial principles and could be examined or refined based on intuition and domain experience. For instance, Kakushadze~\cite{kakushadze2016101} introduces a set of 101 interpretable alpha formulas validated using U.S. market data. The clarity of such formulas facilitates risk auditing, strategy calibration, and theoretical grounding. Nonetheless, hand-crafted alpha factor mining exhibits low development efficiency~\cite{ye2021prediction} and strong subjectivity~\cite{ye2021prediction}. With increasing market complexity driven by structural changes and high-frequency data, manual design of alpha formulas becomes increasingly inadequate, underscoring the need for automated and scalable solutions.

To overcome these limitations, prior studies have explored the automated mining of formulaic alpha factors. Early approaches leveraged heuristic strategies like Genetic Programming (GP)~\cite{zhang2020autoalpha}. GP searches the optimal alpha factor in the solution space represented by nominated symbols including arithmetic operators and financial data features. Within this solution representation, each formulaic alpha is a set of symbols encoded in Reverse Polish Notation (RPN) and is evolved through nature-inspired search operators like mutation, crossover, and selection. Despite the ease of use, they suffer from limited search efficiency due to the ad-hoc search operators, and the lack of theoretical understanding of the search behaviors. This makes them struggle to capture compositional symbolic structures and frequently results in invalid or redundant expressions.

The heavy reliance on the predefined and heuristic rules of GP has motivated a shift toward reinforcement learning (RL) that can learn search behaviors in a data-driven and theoretically-grounded manner. In this paradigm, AlphaGen~\cite{yu2023generating} is the pioneering work, where the alpha factor mining is modeled as a Markov Decision Process (MDP). The RL enables an agent to adaptively explore the symbolic space guided by a neural network-based learnable policy. The agent samples symbols step-by-step and assembles the sample trajectory incrementally as a formulaic expression. The policy is trained using the Information Coefficient (IC), between the real price of the assets and the predicted return of the trajectory, as the reward. This approach enables the constructive generation of interpretable formulaic expressions, whose performance is determined by the trained policy. Follow-up works such as QuantFactor REINFORCE (QFR)~\cite{11024173} successfully contribute to more theoretically stable ways of training the policy.
Unfortunately, RL still faces a critical challenge in solving this MDP. Specifically, the feedback rewards are at the trajectory-level, as rewards are only available after the entire formulaic expression is generated and evaluated. This largely hinders the policy training due to the sparse and delayed rewards, and the resultant low sample efficiency.

In RL, reward shaping has been widely adopted to provide intermediate guidance within a sample trajectory. Techniques like Potential-Based Reward Shaping (PBRS)~\cite{ng1999policy} and the Differential Potential-Based Approach (DPBA)~\cite{wiewiora2003principled} introduce shaping signals without altering the optimal policy. Among them, Reward Shaping from Demonstrations (RSfD)~\cite{brys2015reinforcement} represents a prevalent method to incorporate expert knowledge, using distance-based potential functions derived from expert trajectories. While RSfD provides a practical means to incorporate expert demonstrations via distance-based shaping, its performance hinges on how effectively the shaping signals capture domain-relevant structure. 

In alpha factor mining, such a structure is encoded in symbolic formulas, where existing methods often fail to calculate the correct similarity due to three limitations:
\begin{itemize}
    \item \textbf{Length bias due to discounting:} When the discount factor is less than 1, RSfD tends to prematurely terminate expression generation to avoid discounted rewards, resulting in overly short and under-expressive formulas.
    \item \textbf{Mismatch between syntax and semantics:} RSfD evaluates similarity based on states, making it difficult to distinguish between syntactically similar but semantically different expressions or to recognize semantically equivalent ones which are written differently.
    \item \textbf{Inaccuracy of distance-based metrics:} Tokenized expressions are vectorized for distance computation, yet numerically similar tokens can have vastly different financial meanings. This leads to noisy shaping values and unstable learning.
\end{itemize}
\noindent Consequently, RSfD remains ill-equipped for generation of symbolic expression due to the ineffective demonstration. 

\textit{How can expert-designed formulas be systematically incorporated into the RL framework to improve the training efficiency of alpha mining policy?}
This paper proposes a novel method named Trajectory-Level based Reward Shaping (TLRS). TLRS provides intermediate shaping rewards based on the degree of subsequence-level similarity between partially generated expressions and known expert-designed formulas. By aligning symbolic structures during expression generation, TLRS enables effective knowledge integration while preserving the expressive capacity of RL. Furthermore, its plug-and-play design operates without modifying the policy architecture or introducing additional networks, ensuring seamless compatibility with frameworks such as AlphaGen. The shaping rewards, derived via exact subsequence matching, provide stable and intermediate signals that guide exploration toward structurally meaningful regions in the symbolic space, thus improving convergence speed.


To further stabilize training, a reward centering mechanism is incorporated. By dynamically normalizing the shaping reward relative to the agent’s long-term average return, this strategy alleviates non-stationarity and enhances convergence robustness, especially in volatile learning phases.

Extensive experiments across six major equity indices from Chinese and U.S. stock markets demonstrate that TLRS consistently improves convergence efficiency, robustness, and generalization, outperforming both RL-based and non-RL competitors.

\textbf{The main contributions} of this work are as follows:
\begin{itemize}
    \item A novel framework is established to incorporate expert-designed formulaic alpha factors into RL training via reward shaping.
    \item A shaping mechanism based on exact subsequence alignment is developed to provide structural supervision during the formula generation.
    \item A reward centering strategy is introduced to reduce training variance and improve convergence speed.
\end{itemize}

The remainder of this paper is organized as follows. Section~\ref{Problem Formulation and Preliminaries} introduces the problem formulation and relevant background. Section~\ref{Challenges for RSfD in Alpha Mining} discusses the challenges of applying RSfD to alpha mining. Section~\ref{TLRS} presents the proposed TLRS method in detail. Section~\ref{Numerical Results} reports comprehensive experimental results. Section~\ref{Conclusion} concludes the paper.

This paper uses the following notation: vectors are bold lower case $\mathbf{x}$; matrices are bold upper case $\mathbf{A}$; sets are in calligraphic font $\mathcal{S}$; and scalars are non-bold $\alpha$.

\section{Problem Formulation and Preliminaries}
\label{Problem Formulation and Preliminaries}
This section first introduces the definition of formulaic alpha factors and their RPN sequences. Next, the factor mining MDP modeled by Yu et al. \cite{yu2023generating} are detailed. Finally, to bridge the domain expertise utilization gap in RL-based formulaic alpha mining, an effective technique of reward shaping from demonstrations is introduced.

\subsection{Alpha Factors for Predicting Asset Prices}
\label{Alpha Factors for Predicting Asset Prices} 
Consider a financial market with \( n \) securities observed over \( L \) trading days. Each asset \( i \) on day \( l \in \{1, 2, \cdots, L\} \) is represented as a feature matrix \( \mathbf{X}_{li}\in \mathbb{R}^{m\times d}\), where \( m \) is the number of the adopted raw market features (e.g. open, high, low, close, volume) over the recent \( d \) days. Each element $ \mathbf{x}_{lij} \in \mathbb{R}^{1 \times d} $ represents the $d$-length temporal sequence for the $ j $-th raw market feature, $j = 1,...,m$.
A predictive alpha factor function $ f $ maps the aggregated feature tensor $ \mathbf{X}_l = [\mathbf{X}_{l1}, \mathbf{X}_{l2}, \cdots, \mathbf{X}_{ln}]^\mathsf{T} \in \mathbb{R}^{n \times m \times d} $ to a value vector $ \mathbf{z}_l = f(\mathbf{X}_l)\in \mathbb{R}^{n \times 1} $, where $ \mathbf{z}_l $ contains the $n$ computed alpha values, one for each asset, on the $l$-th day. The complete raw market feature dataset over $L$ days is denoted as $\mathcal{X} = \{\mathbf{X}_l\}^L_{l=1}$.

When a new formulaic alpha factor is generated by the the policy model, it is added to an alpha factors pool $\mathcal{F}=\{f_1,f_2,...,f_K\}$ containing \( K \) validated alpha factors. Following industry practice \cite{RN12}, the linear combination model is adopted to 
predict the asset price as $\mathbf{z}_l^{\prime}=\sum^K_{k=1} w_kf_k(\mathbf{X}_l)$, where $w_k$ is the weight for the $k$-th alpha factor ($k=1,...,K$). In other words, the mining of alpha factors is to search a set of alpha factors and use them in combination.
Each $w_k$ indicates the exposure of the $k$-th factor on the assets. The weights vector $\mathbf{\omega} \in \mathbb{R}^{K\times 1}$ can be optimized through gradient descent by minimizing the prediction error to the ground-truth asset prices $\mathcal{Y} = \{\mathbf{y}_l\}^L_{l=1}$ with \( \mathbf{y}_l \in \mathbb{R}^{n\times 1} \), defined as 
\begin{equation}
    L(\mathbf{\omega})=\frac{1}{L} \sum_{l=1}^{L}\left\|\mathbf{z}_l^{\prime}-\mathbf{y}_l \right\|^{2}.    
    \label{combination model loss}
\end{equation}
Notably, all factors undergo z-score normalization (zero mean, unit maximum) to ensure scale compatibility. When the size of the factor pool grows beyond a preset limit, the factor with the smallest weight will be deleted from the pool.

\subsection{Formulaic Alpha Factors}
Formulaic alpha factors can be encoded as sequences of tokens in Reverse Polish notation (RPN), where each token represents either an operator, a raw price–volume or fundamental feature, a time delta, a constant, or a sequence indicator. The operators include elementary functions that operate on single-day data, known as cross-sectional operators (e.g., Abs($x$) for the absolute value $|x|$, Log($x$) for the natural logarithm of $x$), as well as functions that operate on a series of daily data, known as time-series operators (e.g., Ref($x,l$) for the expression $x$ evaluated at $l$ days before the current day, where $l$ denotes a time token, such as 10d (10 days)). The Begin (BEG) token and Separator (SEP) token of the RPN representation are used to mark the beginning and end of the sequence. Table \ref{All Tokens} illustrates a selection of these tokens as examples.

\begin{table*}[t]
\centering
\caption{Comprehensive Overview of Tokens Used in TLRS}
\begin{tabular}{ccc|ccc|ccc}
\thickhline
Tokens       & Indices & Categories             & Tokens      & Indices & Categories           & Tokens & Indices & Categories         \\ \hline
Abs($x$)       & 0       & Cross-Section Operator & Mad($x,l$)    & 16      & Time-Series Operator & 50     & 32      & Time Span          \\
Log($x$)       & 1       & Cross-Section Operator & Delta($x,l$)  & 17      & Time-Series Operator & -30.0  & 33      & Constant           \\
Add($x,y$)     & 2       & Cross-Section Operator & WMA($x,l$)    & 18      & Time-Series Operator & -10.0  & 34      & Constant           \\
Sub($x,y$)     & 3       & Cross-Section Operator & EMA($x,l$)    & 19      & Time-Series Operator & -5.0   & 35      & Constant           \\
Mul($x,y$)     & 4       & Cross-Section Operator & Cov($x,y,l$)  & 20      & Time-Series Operator & -2.0   & 36      & Constant           \\
Div($x,y$)     & 5       & Cross-Section Operator & Corr($x,y,l$) & 21      & Time-Series Operator & -1.0   & 37      & Constant           \\
Larger($x,y$)  & 6       & Cross-Section Operator & \$open      & 22      & Price Feature        & -0.5   & 38      & Constant           \\
Smaller($x,y$) & 7       & Cross-Section Operator & \$close     & 23      & Price Feature        & -0.01  & 39      & Constant           \\
Ref($x,l$)     & 8       & Time-Series Operator   & \$high      & 24      & Price Feature        & 0.01   & 40      & Constant           \\
Mean($x,l$)    & 9       & Time-Series Operator   & \$low       & 25      & Price Feature        & 0.5    & 41      & Constant           \\
Sum($x,l$)     & 10      & Time-Series Operator   & \$volume    & 26      & Volume Feature       & 1.0    & 42      & Constant           \\
Std($x,l$)     & 11      & Time-Series Operator   & \$vwap      & 27      & Volume-Price Feature & 2.0    & 43      & Constant           \\
Var($x,l$)     & 12      & Time-Series Operator   & 10          & 28      & Time Span            & 5.0    & 44      & Constant           \\
Max($x,l$)     & 13      & Time-Series Operator   & 20          & 29      & Time Span            & 10.0   & 45      & Constant           \\
Min($x,l$)     & 14      & Time-Series Operator   & 30          & 30      & Time Span            & 30.0   & 46      & Constant           \\
Med($x,l$)     & 15      & Time-Series Operator   & 40          & 31      & Time Span            & SEP    & 47      & Sequence Indicator \\ \thickhline
\end{tabular}
\label{All Tokens}
\end{table*}

\begin{figure}[tbh]
\centering
\includegraphics[width=0.47\textwidth]{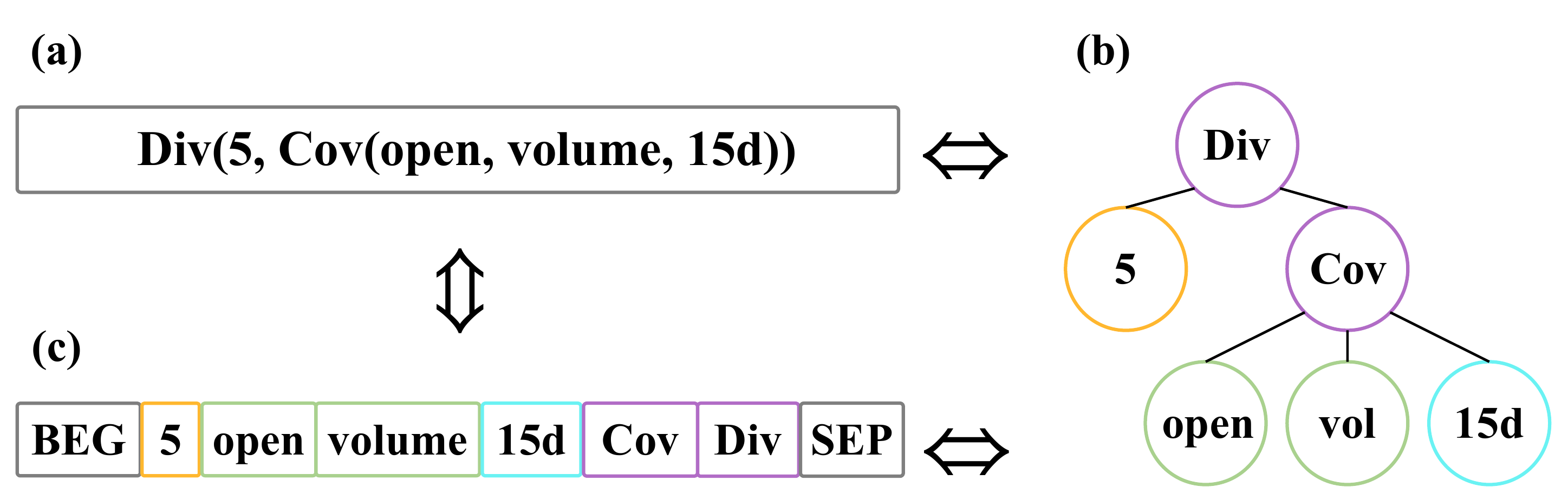}
\caption{Three interchangeable forms of an alpha factor: (a) formulaic expression; (b) tree structure; (c) RPN sequence.}
\label{Figure1}
\end{figure}

Every formulaic alpha factor corresponds to a unique expression tree, with each non-leaf node representing an operator, and the children of a node representing the original volume-price features, fundamental features, time deltas, and constants being operated on. Traversing this tree in post-order yields the RPN representation. Fig. \ref{Figure1} illustrates one such factor alongside its tree and RPN form, and Table \ref{Some Alpha Factor Examples from Alpha 101} presents additional examples drawn from Alpha101 \cite{kakushadze2016101}.

To evaluate an alpha factor’s predictive performance, one commonly computes its Information Coefficient (IC), defined as the Pearson correlation coefficient between the ground-truth asset price \( \mathbf{y}_l \) and the combination factor value \( \mathbf{z}_l^{\prime}\):

\begin{equation} IC\left(\mathbf{z}^{\prime}_l, \mathbf{y}_l\right) = \frac{\text{Cov}(\mathbf{z}^{\prime}_l, \mathbf{y}_l)}{\sigma_{\mathbf{z}^{\prime}_l} \sigma_{\mathbf{y}_l}},
\label{IC defination}
\end{equation}

\noindent where $\mathrm{Cov}(\cdot,\cdot)$ indicates the covariance matrix between two vectors, and $\sigma_{\cdot}$ means the standard deviation of a vector. The IC serves as a critical metric in quantitative finance, where higher values indicate stronger predictive ability. This directly translates to portfolio enhancement through more informed capital allocation: factors with elevated IC enable investors to overweight assets with higher expected returns while underweighting underperformers. Averaging (\ref{IC defination}) over all $L$ trading days gives:
\begin{equation}
  \overline{IC}=\mathbb{E}_{l}\left[IC\left(\mathbf{z}^{\prime}_l, \mathbf{y}_l\right)\right]=\frac{1}{L} \sum^L_{l=1} IC\left(\mathbf{z}^{\prime}_l, \mathbf{y}_l\right).  
  \label{IC_ba}
\end{equation}
\subsection{Mining Formulaic Alpha Factors using RL}
\label{Mining Formulaic Alpha Factors using RL}

The construction of RPN expressions for formulaic alpha factors is formulated as a Markov Decision Process (MDP) \cite{yu2023generating}, formally defined by the tuple $\{\mathcal{S}, \mathcal{A}, P, r, \gamma\}$. Specifically, $\mathcal{A}$ denotes the finite action space, consisting of a finite set of candidate tokens as actions $a$. $\mathcal{S}$ represents the finite state space, where each state at the $t$-th time step corresponds to the sequence of selected tokens, representing the currently generated part of the formulaic expression in RPN, denoted as \(\mathbf{s}_t = \mathbf{a}_{1:t-1} = [a_1, a_2, \cdots, a_{t-1}]^\mathsf{T}\). A parameterized policy $\pi_\mathbf{\theta}:\mathcal{S} \times \mathcal{A} \rightarrow[0,1]$ governs the token selection process, determining action probabilities through sequential sampling $a_{t} \sim \pi_\mathbf{\theta}\left(\cdot \mid \mathbf{a}_{1: t-1}\right)$, where the action $a_t$ is the next token following the currently generated part of the expression $\mathbf{s}_{t}$ in RPN sequence. 

\begin{table*}[tbh]
    \centering
    \caption{Some Alpha Factor Examples from Alpha 101}
    \begin{tabular}{ccc}
    \toprule
    \textbf{Alpha101 Index} & \textbf{Formulaic Expression}                                    & \textbf{RPN Representation}                                     \\ \hline
    Alpha\#6       & Mul(-1, Corr(open, volume, 10d))                        & BEG -1 open volume 10d Corr Mul SEP                    \\
    Alpha\#12      & Mul(Sign(Delta(volume, 1d)), Mul(-1, Delta(close, 1d))) & BEG volume 1d Delta Sign -1 close 1d Delta Mul Mul SEP \\
    Alpha\#41      & Div(Pow(Mul(high, low), 0.5), vwap)                     & BEG high low Mul 0.5 Pow vwap Div SEP                  \\
    Alpha\#101     & Div(Sub(close, open), Add(Sub(high, low), 0.001))       & BEG close open Sub high low Sub 0.001 Add Div SEP      \\ \bottomrule
    \end{tabular}
    \label{Some Alpha Factor Examples from Alpha 101}
\end{table*}

State transition function $P: \mathcal{S} \times \mathcal{A} \times \mathcal{S} \rightarrow[0,1]$ follows the Dirac distribution. Given current state \(\mathbf{a}_{1:t-1}\) and selected action \(a_t\), the subsequent state \(\mathbf{s}_{t+1}\) is uniquely determined as \(\mathbf{s}_{t+1} = \mathbf{a}_{1:t}\) with

\begin{equation}
    P\left(\mathbf{s}_{t+1} \mid \mathbf{a}_{1:t-1}\right)=\left\{\begin{array}{ll}1 & \text { if } \mathbf{s}_{t+1}=\mathbf{a}_{1:t}, \\ 0 & \text { otherwise. }\end{array}\right.
\end{equation}

Valid expression generation initiates with the begin token (BEG), followed by any token selected from $\mathcal{A}$, and terminates (denoted as $t=T$) upon the separator token (SEP) selection or reaching maximum sequence length. To ensure RPN syntax compliance, \cite{yu2023generating} only allow specific actions to be selected in certain states. For more details about these settings, please refer to \cite{yu2023generating}.

The reward function \(r: \mathcal{S} \times \mathcal{A} \rightarrow \mathbb{R}\) assigns values to the state-action pairs and is set to \(r\left(\mathbf{a}_{1: T}\right)=\overline{IC}\) in \cite{yu2023generating}, and $\gamma \in[0,1]$ is the discount rate. It is clear that non-zero rewards are only received at the final $T$-th step, which evaluates the quality of a complete formulaic factor expression, not individual tokens:

\begin{equation}
    r_{t}=\left\{\begin{array}{ll}r\left(\mathbf{a}_{1: T}\right) & \text { if } t = T \\ 0 & \text { otherwise. }\end{array}\right.
    \label{reward function}
\end{equation}

\noindent The objective in this MDP is to learn a policy $\pi_\mathbf{\theta}$ that maximizes the expected cumulative reward over time:

\begin{equation}
    J(\mathbf{\theta})=\mathbb{E}_{\mathbf{a}_{1: T} \sim \pi_{\mathbf{\theta}}}\left[\sum_{t=0}^{T} \gamma^t r_{t}\right] = \mathbb{E}_{\mathbf{a}_{1: T} \sim \pi_{\mathbf{\theta}}}\left[ \gamma^{T} r\left(\mathbf{a}_{1: T}\right)\right].
    \label{original objective function}
\end{equation}

\noindent To ensure $r\left(\mathbf{a}_{1: T}\right)$ is fully propagated back through every intermediate decision, we set the discount factor \(\gamma = 1\). Notably, this MDP only models the formulaic alpha factors generation process, and the environment here specifically refers to the RL environment, rather than the financial market's random behavior. The deterministic feature of transition $P$ is manually designed and holds true regardless of the specific characteristics of the financial market. Expressions that are syntactically correct might still fail to evaluate due to the restrictions imposed by certain operators. For example, the logarithm operator token is not applicable to negative values. Such invalidity cannot be directly omitted. Therefore, these expressions are assigned a reward of $-1$ (the minimum value of IC) to discourage the policy from behaving invalidly.

Based on the factor-mining MDP defined above, Proximal Policy Optimization (PPO) has been used \cite{schulman2017proximal} to optimize the policy $\pi_\mathbf{\theta}\left(a_{t}\mid \mathbf{a}_{1: t-1}\right)$ \cite{yu2023generating}. PPO proposed a clipped objective $L_{surr}\left(\mathbf{\theta}\right)$ as follows:

\begin{align}
        L_{surr}\left(\mathbf{\theta}\right) = \mathbb{E}_{\mathbf{a}_{1:t} \sim \pi_{\mathbf{\theta}_{\text{old}}}} \left.\Bigg[ \sum_{t=1}^{T} A\left( \mathbf{a}_{1:t} \right) \min \left.\Big( \psi\left( \mathbf{a}_{1:t} \right), \right. \right. \nonumber \\ 
        \left. \left. \operatorname{clip}\left( \psi\left( \mathbf{a}_{1:t} \right), 1-\delta, 1+\delta \right) \right.\Big) \right.\Bigg],
\end{align}
\noindent where ratio $\psi\left(\mathbf{a}_{1:t}\right) = \pi_{\mathbf{\theta}}\left(a_{t} \mid \mathbf{a}_{1:t-1}\right) / \pi_{\mathbf{\theta}_{\text {old }}}\left(a_{t} \mid \mathbf{a}_{1:t-1}\right)$ is the importance weight, and \(A\left(\mathbf{s}_{t}, a_{t}\right) = Q_{\pi_{\mathbf{\theta}}}(a_t, \mathbf{a}_{1:t-1}) - V_{\pi_{\mathbf{\theta}}}(\mathbf{a}_{1:t-1})\) is an estimator of the advantage
function at timestep $t$. $Q_{\pi_{\mathbf{\theta}}}(a_t, \mathbf{a}_{1:t-1})$ is the state-action value function, and $V_{\pi_{\mathbf{\theta}}}(\mathbf{a}_{1:t-1})$ is the state value function.

\subsection{Reward Shaping from Demonstrations}

In MDPs with trajectory-level rewards, the initial training lacks any prior knowledge, causing the initial policy to randomly explore different state-action pairs. Only after gathering enough transitions and rewards can the agent start favoring actions that perform better. Reward shaping modifies the original reward with a potential function and provides the capability to address the above cold-start issue in training. The additive form, which is the most general form of reward shaping, is considered. Formally, this can be defined as $r_t^{\prime}=r_t+f_t$, where $r_t$ is the original reward, $f_t$ is the shaping reward. $f_t$ enriches the sparse trajectory-level reward signals, providing useful gradients to the agent. Early work of reward shaping\cite{dorigo1994robot} focuses on designing the shaping reward $f_t$, but ignores that the shaping rewards may change the optimal policy. Potential-based reward shaping (PBRS) is the first approach which guarantees the so-called optimal policy invariance property\cite{ng1999policy}. Specifically, PBRS defines $f_t$ as the difference of potential values: 
\begin{equation}
f\left(\mathbf{s}_t, a_t, \mathbf{s}_{t+1}\right)=\gamma \Phi\left(\mathbf{s}_{t+1}\right)-\Phi(\mathbf{s}_t),
\label{shaping function}
\end{equation}

\noindent where $\Phi: \mathcal{S} \rightarrow \mathbb{R}$ is a potential function which gives hints on states. 

In MDPs where it is non-trivial to define an effective potential function, demonstrations can be provided to prevent the early unstable training of RL\cite{xiao2022cold}. To enriches the sparse trajectory-level reward signals, the agent is given a series of human expert demonstrations, typically in the form of state-action pairs $\left\{\left(\mathbf{s}_t^e, a_t^e\right)\right\}_{t=0}^n$. This method is referred as Reward Shaping from Demonstrations (RSfD)\cite{brys2015reinforcement}. RSfD encodes the demonstrations in the learning process as a potential-based reward shaping function. Specifically, RSfD defines a state-similarity metric between the agent’s trajectories and expert demonstrations, then find the sample that yields the highest similarity $\Phi(\mathbf{s}_t)=\max g\left(\mathbf{s}_t, \mathbf{s}^e_t\right)$, where $g\left(\mathbf{s}_t, \mathbf{s}^e_t\right)$ refers to a similarity metric normally involving distance-based or distribution-based methods\cite{brys2015reinforcement}. Common distance-based metrics (e.g., Euclidean\cite{sutton2018reinforcement}, Manhattan\cite{bellemare2013arcade}, Cosine\cite{argall2009survey}, Mahalanobis\cite{abbeel2004apprenticeship}, and Hamming distances\cite{nouri2008multi}) measure how far apart two states are, while probability distribution comparisons (e.g., Kullback-Leibler Divergence\cite{ziebart2008maximum} and Earth Mover’s Distance\cite{zhao2024mimic}) capture how state distributions differ.

\section{Challenges for RSfD in Alpha Mining}
\label{Challenges for RSfD in Alpha Mining}
This section discusses three main problems arising from adopting RSfD in formulaic alpha factor mining. A primary concern is that a discount factor other than 1 can cause premature termination of the factor mining process, which in turn affects the validity of the factor expressions. Additionally, relying solely on syntactic structure makes it challenging to accurately capture the true semantics of factor expressions in different states. Finally, distance-based similarity metrics possess inherent limitations in evaluating the similarity between states.
\subsection{Fully Long-term Reward Orientation}
If the discount factor $\gamma$ is not 1, the factor-mining MDP may not function as expected. In this setting, the agent can choose to use the SEP action to terminate the MDP and receive a non-zero reward (defined as $\overline{IC}$ in (\ref{IC_ba})). Consequently, the term $\gamma^{T}$ in the optimization objective (\ref{original objective function}) depends on the policy parameter $\mathbf{\theta}$. Whenever $\gamma < 1$, the policy tends to end tasks prematurely (opting for shorter-length factors) to maximize (\ref{original objective function}) due to the term $\gamma^{T}$. This leads to several issues: the agent may select shorter-length factors that fail to represent market features adequately; many truncated expressions appear early in exploration, wasting computation and delaying the discovery of effective factors; and the term $\gamma^{T}$ depends on factor length, introducing higher variance into the objective function and destabilizing convergence. A theoretical proof is provided in Proposition 1 of Section \ref{The Theoretical Analysis}, demonstrating that when the discount factor $\gamma$ is not 1, the agent tends to terminate the MDP earlier and generate shorter-length factors.

\subsection{Imbalance Among States}
The imbalance between different states in the factor-mining MDP is also an important issue. As described in Section \ref{Mining Formulaic Alpha Factors using RL}, the states of the factor-mining MDP consist of tokens already generated, which can be converted into formulaic alpha factors through RPN. Measuring state-similarity between the agent’s trajectories and expert demonstrations means comparing these formulaic alpha factors. However, semantically equivalent factors may differ in syntax (e.g., $\text{close}^2 + 2\text{close} + 1$ vs. $(\text{close} + 1)^2$), while syntactically similar factors may differ in semantics (e.g., $(\text{close} + \text{high})*5$ vs. $\text{close} + \text{high}*5$). Furthermore, longer factors generally have a stronger representational capacity, making it impossible to directly compare factors of shorter lengths. These two aspects make comparing these formulaic factors non-trivial. An ideal solution uses a representation model that gives algebraic expressions with similar semantics a similar continuous representation even when syntax differs\cite{allamanis2017learning}. A simpler yet effective solution is proposed in Section \ref{The Proposed Algorithm}.

\subsection{Limitations of Distance-Based Similarity Metrics}
Distance-based metrics cannot compute $g\left(\mathbf{s}_t, \mathbf{s}^e_t\right)$ measuring the similarity between the current state and an expert demonstration state. Specifically, RPN expressions are vectorized by assigning each token a numerical index from a dictionary. As shown in Table \ref{All Tokens}, tokens “open” and “close” may have a small numerical difference, yet their market significance differs greatly. Because numerical differences in the RPN vector do not reflect true semantic differences, distance-based similarity metrics may produce inaccurate and high-variance shaping rewards that harm the agent’s learning process.

\begin{figure*}[!t]
\centering
\includegraphics[width=0.8\textwidth]{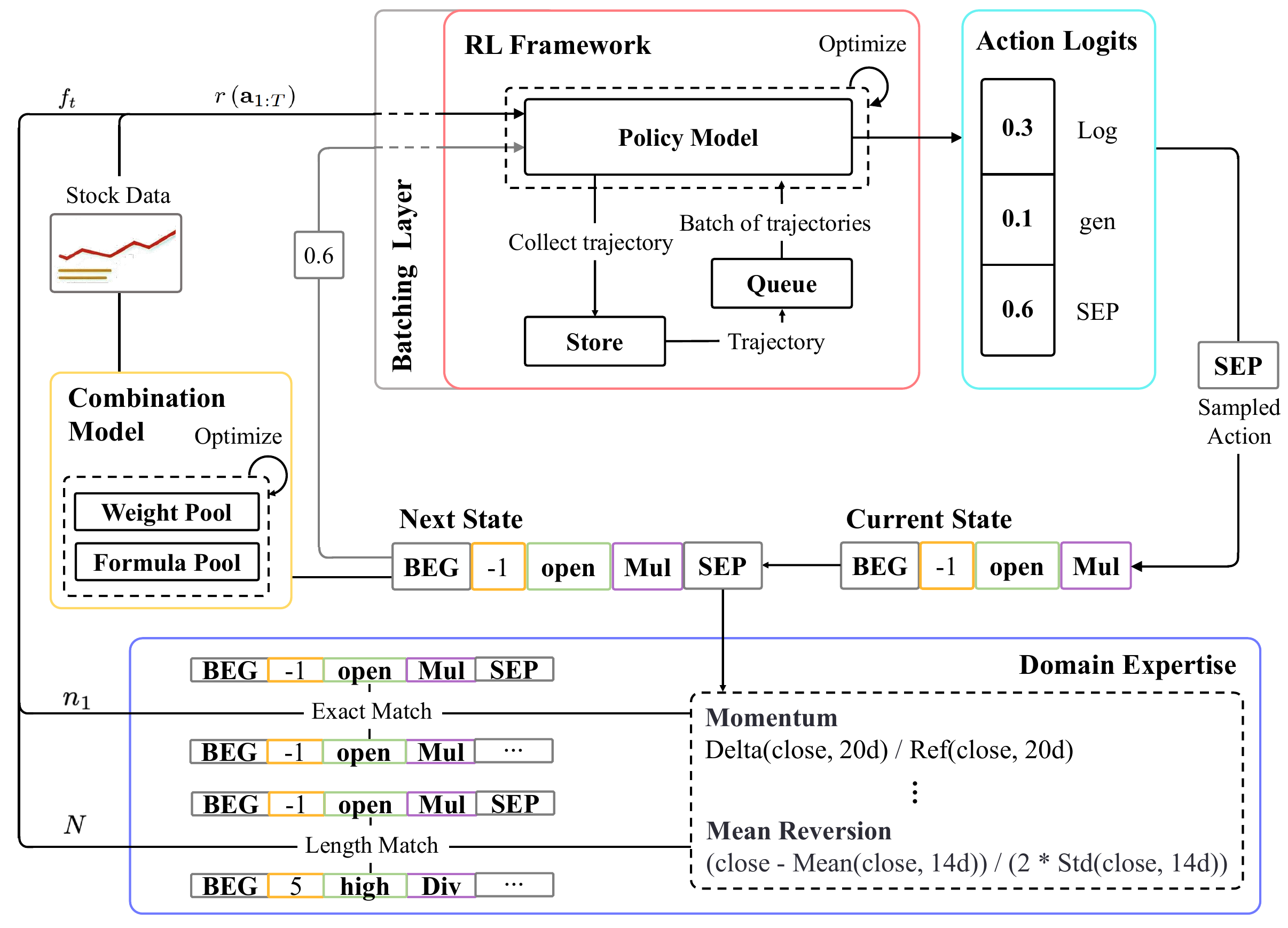}
\caption{The detailed pipeline for the proposed TLRS algorithm. }
\label{Figure2}
\end{figure*}

\section{Trajectory Level-Based Reward Shaping}
\label{TLRS}
To address the limitations of traditional methods in effectively incorporating domain expertise and to overcome the challenges in reward shaping for alpha mining detailed in Section \ref{Challenges for RSfD in Alpha Mining}, this section introduces Trajectory-level Reward Shaping (TLRS). TLRS, a simple yet effective approach, is specifically designed for factor-mining MDPs where the discount factor $\gamma$ must be 1. It proposes a novel similarity metric based on exact subsequence matching between the agent's trajectories and expert demonstrations, tackling issues of semantic ambiguity and the shortcomings of distance-based metrics. Furthermore, a reward centering method is integrated to enhance training stability by reducing reward variance. Theoretical analyses support optimal policy invariance and computational efficiency under TLRS, as well as the benefits of reward centering.
\subsection{The Proposed Algorithm}
\label{The Proposed Algorithm}
TLRS establishes a noval similarity metric between agent trajectories and expert demonstrations through subsequence matching, defined as $n_{1,t}/N_{t}$, where $n_{1,t}$ denotes the number of expert demonstration subsequences exactly matching the current generated partial sequence $\mathbf{s}_t$. $N_t$ represents the total number of subsequences in expert demonstrations with length $t$. The term $n_{1,t}/N_{t}$ calculates the exact match ratio for generated partial sequences $\mathbf{s}_t$ with length $t$. The shaping reward \( f_t \) is computed as the temporal difference of exact match ratios between consecutive steps in the formulaic factor mining MDP:

\begin{equation}
f_t = \delta(\mathbf{s}_t, \mathbf{s}_{t+1}) = \frac{n_{1,t+1}}{N_{t+1}} - \frac{n_{1,t}}{N_t}.
\label{TLRS_eq}
\end{equation}

\noindent This formulation ensures that \( f_t \) captures the incremental alignment between the agent's trajectory and expert demonstrations through progressive subsequence matching. The detailed calculation pipeline is depicted in Fig. \ref{Figure2}.

If the term $n_{1,t}/N_{t}$ is defined as the potential function $\Phi(\mathbf{s}_{t+1})$, then (\ref{TLRS_eq}) shares the same mathematical form with PBRS:

\begin{equation}
f_t = \gamma\Phi(\mathbf{s}_{t+1}) - \Phi(\mathbf{s}_t),
\label{f_t_phi}
\end{equation}

\noindent where $\gamma=1$ as required by the factor-mining MDP. In Proposition2 of Section \ref{The Theoretical Analysis}, we provide theoretical guarantees regarding the invariance of the optimal objective function, following the seminal work\cite{ng1999policy}.

Although (\ref{TLRS_eq}) shares mathematical form with PBRS, they embody completely different underlying ideas. TLRS defines the state's potential $\Phi(\mathbf{s}_t)$ as a ratio. The numerator counts how many expert demonstration subsequences exactly match the current policy's partial sequence of length $t$. The denominator is the total number of expert demonstration subsequences of length $t$. Assuming the expert factor dataset is generated by a model sharing the same network architecture as the current policy model being optimized, with weights denoted by \(\mathbf{\theta}_e\), while the current policy model has weights \(\mathbf{\theta}\). TLRS measures the gap between \(\mathbf{\theta}_e\) and \(\mathbf{\theta}\) through \( f_t \) defined in (\ref{TLRS_eq}). As \(\mathbf{\theta}\) approaches \(\mathbf{\theta}_e\), the generated sequences more closely align with expert demonstrations, leading to increased matching ratios \(n_{1,t+1}/N_{t+1}\) and \(n_{1,t}/N_t\) at all time steps. Because matching longer sequences is more challenging (short sequences are easy to match, while longer sequences are nearly impossible to match through random generation—note that for a policy sequence of length \(t+1\) to match an expert subsequence of length \(t+1\), their subsequences of length \(t\) must be identical), the matching ratio at time step \(t+1\) is usually low during the early stages of training. However, its growth rate surpasses that of timestep $t$ as learning progresses. Therefore, the matching increments $f_t$ increase as \(\mathbf{\theta}\)  approaches \(\mathbf{\theta}_e\), thereby producing stronger learning signals to reinforce policy-expert behavior consistency. In contrast, PBRS typically employs heuristic-driven potential functions that correlate with state value estimations, using $\gamma\Phi(\mathbf{s}_{t+1}) - \Phi(\mathbf{s}_t)$ to encourage transitions toward high-potential states through unidirectional incentives. 

Shaped by $f_t$, the reward function (\ref{reward function}) becomes:
\begin{equation}
    r^\prime_{t}=\left\{\begin{array}{ll}f_t & \text { if } t \neq T \\ f_t +r\left(\mathbf{a}_{1: T}\right) & \text { otherwise. }\end{array}\right.
    \label{Shaped Reward}
\end{equation}
\noindent Notably, TLRS is exclusively applicable to MDPs with $\gamma = 1$. If $\gamma \neq 1$, then $f_t$ no longer represents the matching increments, and simultaneously, $\Phi(\mathbf{s}_{t+1})$ will be compressed, causing the matching signal  $\Phi(\mathbf{s}_{t})$ to dominate. Moreover, regarding the semantic consistency of formulas, there are two cases: one in which both semantics and syntax are identical—in this case, it is enough to ensure that the two vectorized formula are completely identical, as stated in (\ref{TLRS_eq}); and the other where only the semantics are consistent while the syntax differs. We have proven that the error caused by ignoring the second situation is bounded in Proposition 3 of Section \ref{The Theoretical Analysis}. Therefore, directly using the ratio of the expert demonstration subsequences that exactly match the generated sequence is acceptable, and it greatly reduces the computational complexity.

\begin{algorithm}[b]  
	\caption{TLRS}
	\LinesNumbered 
	\KwIn{Real asset price dataset $\mathcal{Y}$, Real asset feature dataset $\mathcal{X}=\{\mathbf{X}^{\prime}_l\}$, Initial policy weight $\mathbf{\theta}$, Initial combination model weight $\mathbf{\omega}$, Discount factor $\gamma$.}
	\KwOut{Formulaic alpha factors generator $\pi_\mathbf{\theta}$.}
	\While{not converged}{
        Construct a factor $f_n$ with $\pi_{\mathbf{\theta}}\left(\cdot \mid \mathbf{a}_{1: t-1}\right)$\;
        Compute factor values $\{\mathbf{z}_{n,l}\}=\{f_n(\mathbf{X}_l)\}$\;
        Compute $\{\mathbf{z}^{\prime}_{n,l}\}$ of the factor with $\mathbf{\omega}$\;
        Compute $r\left(\mathbf{a}_{1: T}\right)$ with the shaped reward and $\{\mathbf{z}^{\prime}_{n,l}\}$ via Eq. (\ref{Shaped Reward})\;
        Compute the online estimation \(\bar{r}_{t}\) of the average reward \(r(\pi_{\mathbf{\theta}})\) and update $\mathbf{\theta}$ via Eq. (\ref{centered TD})\;
        Update $\mathbf{\omega}$ via optimizing Eq. (\ref{combination model loss})\;	
	}
\label{ag1}
\end{algorithm}

\subsection{Reward Centering}
Although TLRS dramatically accelerates convergence compared to other RL baselines, its reward curve still exhibits large oscillations during training. We hypothesize that the variations in shaping rewards contribute to this instability. Inspired by Blackwell's Laurent decomposition theory\cite{naik2024reward}, we decompose the value function in factor-mining MDP as: 

\begin{equation}
V_{\pi_{\mathbf{\theta}}}(\mathbf{s}) = \mathbb{E}_{\pi_{\mathbf{\theta}}}\left[\sum_{t=0}^{T}  r(\pi_{\mathbf{\theta}}) \mid \mathbf{s}_0 = \mathbf{s}\right] + \tilde{V}_{\pi_{\mathbf{\theta}}}(\mathbf{s}),
\label{decomposition}
\end{equation}
\noindent where \(r(\pi_{\mathbf{\theta}})\) is the state-independent average step reward depend on \(\pi_{\mathbf{\theta}}\), defined as follows:
\begin{equation}
    r(\pi_{\mathbf{\theta}}) \doteq \lim_{T \to \infty} \frac{1}{T} \mathbb{E}_{\pi_{\mathbf{\theta}}} \left[ \sum_{t=0}^{T} r_t \right].
\end{equation}

\noindent The first term \(\mathbb{E}_{\pi_{\mathbf{\theta}}}\left[\sum_{t=0}^{T}  r(\pi_{\mathbf{\theta}}) \mid \mathbf{s}_0 = \mathbf{s}\right]\) is a state-independent constant, and \(\tilde{V}_{\pi_{\mathbf{\theta}}}(\mathbf{s})\) is the differential value of state \(\mathbf{s}\)\cite{naik2024reward}, defined as follows:  
\begin{equation}
    \tilde{V}_{\pi_{\mathbf{\theta}}}(\mathbf{s}) \doteq \mathbb{E}_{\pi_{\mathbf{\theta}}}\left[\sum_{t=0}^{T}  r_{t} - r(\pi_{\mathbf{\theta}}) \mid \mathbf{s}_0 = \mathbf{s}\right].
\end{equation}

\noindent A detailed derivation of this decomposition is provided in Proposition 4 of Section \ref{The Theoretical Analysis}. Based on (\ref{decomposition}), a reward centering mechanism tailored for the factor-mining MDP can be established. The average reward \(r(\pi_{\mathbf{\theta}})\) is dynamically tracked through online estimation \(\bar{r}_{t+1} \doteq  \bar{r}_t + \beta(r_{t+1}-\bar{r}_t)\). This update leads to an unbiased estimate of the average reward \(\bar{r}_{t} \approx r(\pi_{\mathbf{\theta}})\)\cite{naik2024reward}. The TD update can then be reconstructed as:
\begin{align}
    \tilde{V}_{\pi_{\mathbf{\theta}}}(\mathbf{a}_{1:t-1}) \leftarrow \tilde{V}_{\pi_{\mathbf{\theta}}}(\mathbf{a}_{1:t-1})+\alpha\bigg[\left(r_{t+1}-\bar{r}_t\right) \nonumber\\
    +\tilde{V}_{\pi_{\mathbf{\theta}}}(\mathbf{a}_{1:t})-\tilde{V}_{\pi_{\mathbf{\theta}}}(\mathbf{a}_{1:t-1})\bigg].
    \label{centered TD}
\end{align}

\noindent Mathematically, this shifts the learning objective from the original value function $V_{\pi_{\mathbf{\theta}}}(\mathbf{a}_{1:t-1})$ to the differential value function $\tilde{V}_{\pi_{\mathbf{\theta}}}(\mathbf{a}_{1:t-1})$. It neutralizes the explosive state-independent constant term \(\mathbb{E}_{\pi_{\mathbf{\theta}}}\left[\sum_{t=0}^{T}  r(\pi_{\mathbf{\theta}}) \mid \mathbf{s}_0 = \mathbf{a}_{1:t-1}\right]\) in the decomposition. As a result, the value function approximator no longer needs to fit the bias term, allowing it to focus more effectively on the relationship between state features and the differential value of state \(\tilde{V}_{\pi_{\mathbf{\theta}}}(\mathbf{a}_{1:t-1})\). The algorithmic steps are given in Algorithm 1. 

\subsection{The Theoretical Analyses}
\label{The Theoretical Analysis}
\textbf{Proposition 1.} {Consider an MDP with trajectory-level rewards and a termination action, 
Let $r_{T}(\mathbf{\theta})=\mathbb{E}_{\mathbf{s}_T}\left[r(\mathbf{s}_T)\right]$ denote the average reward for complete trajectories of length $T \in \{1,\cdots, T_{max}\}$, where $T_{max}$ denotes the maximum permissible trajectory length. For $\gamma<1$, if any pair of consecutive trajectory lengths satisfies $r_{T}>\gamma r_{T+1}$, the agent strictly prefers sequences of length $T$ over $T+1$, despite potentially lower rewards ($r_{T}< r_{T+1}$). When $\gamma = 1$, the agent exhibits no inherent preference between trajectories of different lengths, and purely seeks larger rewards.}

\begin{proof}
Since the reward is only given at the final step of each trajectory, the objective function can be expressed as the expectation value over all possible complete expressions $\mathbf{s}_T$:

\begin{align}
J(\mathbf{\theta}) &= \mathbb{E}_{\mathbf{s}_T}\left[\gamma^{T}r(\mathbf{s}_T)\right] \nonumber\\ 
&=\mathbb{E}_T \left[\gamma^{T}\cdot\mathbb{E}_{\mathbf{s}_T}\left[r(\mathbf{s}_T)\right] \right] \nonumber\\
&=\mathbb{E}_T \left[\gamma^{T}\cdot r_{T}(\mathbf{\theta}) \right] \nonumber\\
&=\sum_{T=1}^{T_{max}}p_T(\mathbf{\theta})\gamma^{T}r_{T}(\mathbf{\theta}), \nonumber
\end{align}

\noindent where $p_T(\mathbf{\theta})$ represents the probability of the model generating a complete expression of length $T$, $r_{T}(\mathbf{\theta})=\mathbb{E}_{\mathbf{s}_T}\left[r(\mathbf{s}_T)\right]$ denotes the average reward for complete expressions of length $T$. Then the gradient of the objective function is derived as:

\begin{align}
&\frac{\partial J(\mathbf{\theta})}{\partial \mathbf{\theta} } =\sum_{T=1}^{T_{max}}p_{T}\cdot\gamma^T\frac{\partial r_{T}}{\partial \mathbf{\theta}}+\sum_{T=1}^{T_{max}}\frac{\partial p_{T}}{\partial \mathbf{\theta}}\gamma^T  r_{T}. \nonumber
\end{align}

\noindent Let $P_T$ denote the probability of generating expressions with length $\leq T$, and thus  $p_T=P_T-P_{T-1}$ ($T\geq1$) represents the probability of generating expressions of exactly length $T$. The second gradient term can be rewritten as:

\begin{align}
\sum_{T=1}^{T_{max}}\frac{\partial p_{T}}{\partial \mathbf{\theta}}\gamma^T  r_{T}&=\sum_{T=1}^{T_{max}}\frac{\partial P_{T}}{\partial \mathbf{\theta}}\gamma^T  r_{T}-\sum_{T=1}^{T_{max}}\frac{\partial P_{T-1}}{\partial \mathbf{\theta}}\gamma^T  r_{T} \nonumber \\
&=\sum_{T=1}^{T_{max}}\frac{\partial P_{T}}{\partial \mathbf{\theta}}\gamma^T  r_{T}-\sum_{T=0}^{T_{max}-1}\frac{\partial P_{T}}{\partial \mathbf{\theta}}\gamma^{T+1}  r_{T+1} \nonumber \\
&=\sum_{T=1}^{T_{max}}\frac{\partial P_{T}}{\partial \mathbf{\theta}}\gamma^T  r_{T}-\sum_{T=1}^{T_{max}}\frac{\partial P_{T}}{\partial \mathbf{\theta}}\gamma^{T+1}  r_{T+1}  \label{prop1_switchT}\\
&=\sum_{T=1}^{T_{max}}\frac{\partial P_{T}}{\partial \mathbf{\theta}}\gamma^T(r_{T}-\gamma r_{T+1}), \nonumber
\end{align}

\noindent where (\ref{prop1_switchT}) follows from $P_0 = 0$ and $P_{T_{max}}=1$, leading to $\frac{\partial P_{0}}{\partial \mathbf{\theta}}=\frac{\partial P_{T_{max}}}{\partial \mathbf{\theta}}=0$. The total gradient thus becomes:

\begin{align}
\frac{\partial J(\mathbf{\theta})}{\partial \mathbf{\theta} }
&=\sum_{T=1}^{T_{max}}p_{T}\cdot\gamma^T\frac{\partial r_{T}}{\partial \mathbf{\theta}}+\sum_{T=1}^{T_{max}}\frac{\partial P_{T}}{\partial \mathbf{\theta}}\gamma^T(r_{T}-\gamma r_{T+1}). \nonumber 
\end{align}

\noindent The total gradient comprises two components: the first term relates to gradients with respect to rewards, while the second term relates to gradients on $P_{T}$. Notably, $P_{T}$ also represents the model's likelihood of terminating generation by selecting 'SEP' token at $T+1$ position. Consequently, this gradient term optimizes the objective function through adjusting trajectory length $T$.

That is, the second gradient term exhibits model's preference on trajectory length. When $\gamma$ is set to less than one, if $r_{T}<r_{T+1}$ but $r_{T}>\gamma r_{T+1}$, the gradient direction aligns with $\frac{\partial P_{T}}{\partial \mathbf{\theta}}$, causing the model to favor shorter expressions even when they yield lower rewards. This indicates that the objective function introduces a bias toward shorter expressions. Setting $\gamma=1$ eliminates this bias, thereby avoiding the unintended preference for shorter trajectories.

\end{proof}

\textbf{Proposition 2.} \textit{Let $M = \{\mathcal{S}, \mathcal{A}, P, r, \gamma\}$ be the original MDP with the reward function $r_t$, and $M^{\prime} = \{\mathcal{S}, \mathcal{A}, P, r^\prime, \gamma\}$ be the shaped MDP with the shaped reward function $r^{\prime}_t=f_t+r_t$, where $f_t$ defined in (\ref{f_t_phi}). Let $\pi_M^*$ and $\pi_{M^{\prime}}^*$, be the optimal policies in $M$ and $M^{\prime}$, respectively. Then, $\pi_{M^{\prime}}^*$ is consistent with $\pi_M^*$.} 

\begin{proof}
The optimal Q-function in $M$ should be equal to the expectation of long-term cumulative reward as:
$$
Q_M^*(s, a)=\mathbb{E}\left[\sum_{t=0}^{T} \gamma^t r_t \mid \mathbf{s}_0 = s, a_0 = a\right].
$$
Likewise, the optimal Q-function in $M^{\prime}$ can be denoted as:
$$
\begin{aligned}
Q_{M^{\prime}}^*(s, a) & =\mathbb{E}\left[\sum_{t=0}^{T} \gamma^t r_t^{\prime}  \mid \mathbf{s}_0 = s, a_0 = a\right] \\
& =\mathbb{E}\left[\sum_{t=0}^{T} \gamma^t\left(r_t+f_t\right) \mid \mathbf{s}_0 = s, a_0 = a \right] .
\end{aligned}
$$

\noindent According to (\ref{f_t_phi}), we have:

$$
\begin{aligned}
Q_{M^{\prime}}^*(s, a)= & \mathbb{E}\left[\sum _ { t = 0 } ^ { T } \gamma ^ { t } \left(r_t+\gamma\Phi(\mathbf{s}_{t+1}) - \Phi(\mathbf{s}_t)\right)\right] \\
= & \mathbb{E}\left[\sum_{t=0}^{T} \gamma^t r_t\right]+\mathbb{E}\left[\sum_{t=1}^{T} \gamma^t \Phi_t\left(\mathbf{s}_t\right)\right]\\
& - \mathbb{E}\left[\sum_{t=0}^{T} \gamma^t \Phi_t\left(\mathbf{s}_t\right)\right] \\
= & \mathbb{E}\left[\sum_{t=0}^{T} \gamma^t r_t\right]-\Phi_0\left(\mathbf{s}_0\right).
\end{aligned}
$$

\noindent Thus, we have $Q_{M^{\prime}}^*(s, a)=Q_M^*(s, a)-\Phi_0\left(\mathbf{s}_0\right)$, where $\Phi_0\left(\mathbf{s}_0\right)$ denotes the initial value of the potential function. The policy is obtained by maximizing the value of Q-function, and hence the optimal policy in $M^{\prime}$ can be expressed as
$$
\begin{aligned}
\pi_{M^{\prime}}^* & =\underset{\pi}{\arg \max } Q_{M^{\prime}}^*(\mathbf{s}, a) \\
& =\underset{\pi}{\arg \max }\left[Q_M^*(\mathbf{s}, a)-\Phi_0\left(\mathbf{s}_0\right)\right]\\
& =\underset{\pi}{\arg \max }Q_M^*(\mathbf{s}, a).
\end{aligned}
$$

\noindent Therefore, $\pi_{M^{\prime}}^*=\arg \max _{\pi} Q_{M}^*(s, a)=\pi_M^*$, which demonstrates that $\pi_{M^{\prime}}^*$ is consistent with $\pi_M^*$. This completes the proof.
\end{proof}

\textbf{Proposition 3.} \textit{Consider a randomly generated sequence being compared with a set of expert demonstration subsequences. Let $f_t$ denote the shaping reward where the potential function is defined as the ratio of expert demonstration subsequences sharing identical semantics and syntax with the generated sequence, and let $f_t^*$ denote the ideal shaping reward that accounts solely for semantics, disregarding syntax consistency. The error induced by replacing $f_t^*$ with $f_t$ is bounded by $\epsilon_t=\left|\frac{
f_t^*-f_t
}{f_t^*}\right|<\left(\frac{t}{m}\right)^2k$, where $t$ is the sequence length, $m$ is the token vocabulary size, and $k$ is the number of operator types which might induce syntax inconsistency. This error vanishes under $m^2\gg t^2k$.} 

\begin{proof}
Due to the randomness of sequence generation process, we analyze the counts of matching sequence-expert subsequence pairs, instead of matching expertise subsequences for a given sequence. A possible matching pair might arise from two cases: 1) identical semantics and identical syntax; 2) identical syntax despite differing semantics. Let $N_t$ denote the total number of sequence-expertise subsequence pairs of length \(t\), with $n_{1,t}$ and $n_{2,t}$ denoting the number of matching pairs from case 1) and 2), respectively.

To calculate $n_{1,t}$, we first consider the total number of possible sequences. Given a sequence length of $t$ and a token vocabulary size of $m$, there are $m^t$ possible unique sequences. For case 1), where a pair must have identical semantics and syntax, each of these $m^t$ sequences is uniquely paired with an identical copy of itself. Therefore, the number of matching pairs for case 1) is: 
\begin{equation*}
    n_{1,t}=m^t.
\end{equation*}
\noindent Before calculating $n_{2,t}$, we note that case 2) occurs when expressions admit equivalent transformations, causing the two sequences to differ in at least two tokens. For example, syntax variations might arise from: commutativity ($x + y = y + x$, related to a 2-token difference in Reverse Polish Notation) or logarithmic identities ($\log(a*x) = \log(x) + \log(a)$, related to a 3-token difference). Each case contributes to $n_{2,t}$ by a term of $2m^{t-d}\binom{t}{d}\leq2m^{t-2}\frac{t^d}{d!}$, where $d$ is the number of different tokens. Since changing a single token always alters the expression's semantics, any pair of equivalent sequences must differ in at least two tokens. Given the number of different tokens $d\geq2$ and commonly $m>t$, each contributing term is bounded by $2m^{t-d}\frac{t^d}{d!}\leq m^{t-2}t^2$. $n_{2,t}$ can be then approximated by:
$$
    n_{2,t}\approx m^{t-2}\cdot t^2k,
$$

\noindent where $k$ is the number of operator types which might induce syntax inconsistency. The ratio of $n_{2,t}$ to $n_{1,t}$ is:
 \begin{equation}
    \frac{n_{2,t}}{n_{1,t}}\approx \left(\frac{t}{m}\right)^2k. \nonumber
\end{equation}

\noindent From \ref{TLRS_eq} and \ref{f_t_phi}, the potential functions are $\Phi_t=n_{1,t}/N_t$. Accounting solely for semantics and disregarding syntax consistency, the ideal potential functions are $\Phi^*_t=(n_{1,t}+n_{2,t})/N_t$. Their corresponding shaping rewards are computed from $f_t=\Phi_{t+1}-\Phi_{t}$ and $f_t^*=\Phi_{t+1}^*-\Phi_{t}^*$. The difference between $f_t^*$ and $f_t$ satisfies:
\begin{align}
    f^*_{t}-f_{t}=&\left(\frac{n_{1,t+1}+n_{2,t+1}}{N_{t+1}}-\frac{n_{1,t}+n_{2,t}}{N_{t}}\right) \nonumber\\&-\left(\frac{n_{1,t+1}}{N_{t+1}}-\frac{n_{1,t}}{N_{t}}\right) \nonumber\\
=&\frac{n_{2,t+1}}{N_{t+1}}-\frac{n_{2,t}}{N_{t}} \nonumber\\
=&\frac{n_{2,t+1}}{n_{1,t+1}+n_{2,t+1}}\Phi_{t+1}^*-\frac{n_{2,t}}{n_{1,t}+n_{2,t}}\Phi_{t}^* \nonumber\\
=&\frac{\left( \frac{t+1}{m} \right)^2k}{1+\left( \frac{t+1}{m} \right)^2k}\Phi_{t+1}^*-\frac{\left( \frac{t}{m} \right)^2k}{1+\left( \frac{t}{m} \right)^2k}\Phi_{t}^* \nonumber\\
\geq &\frac{\left( \frac{t}{m} \right)^2k}{1+\left( \frac{t}{m} \right)^2k}\left( \Phi_{t+1}^*-\Phi_{t}^* \right) \nonumber\\
=&\frac{\left( \frac{t}{m} \right)^2k}{1+\left( \frac{t}{m} \right)^2k}f^*_{t}. \nonumber
\end{align}

\noindent Because matching longer sequences is more challenging (short sequences are
easy to match, while longer sequences are nearly impossible to match through random generation—note that for a policy sequence of length $t + 1$ to match an expert subsequence of length $t + 1$, their subsequences of length $t$ must be identical), the matching ratio $\Phi^*_{t}$ decreases with sequence length $t$ increases, implying that $f^*_{t}=\Phi^*_{t+1}-\Phi^*_{t}$ is always negative. Similarly, the absolute error $f^*_{t}-f_{t}=\frac{n_{2,t+1}}{N_{t+1}}-\frac{n_{2,t}}{N_{t}}$ is also negative. Therefore, the relative error is bounded by:
\begin{equation}
    \epsilon_t=\left|\frac{
f_t^*-f_t
}{f_t^*}\right|\leq\frac{\left( \frac{t}{m} \right)^2k}{1+\left( \frac{t}{m} \right)^2k}<\left(\frac{t}{m}\right)^2k. \nonumber
\end{equation}

\noindent This error vanishes under condition $m^2\gg t^2k$.

\end{proof}

\textbf{Proposition 4.} \textit{Consider an ergodic MDP with finite state and action spaces. Subtracting the average policy reward \(r(\pi_\mathbf{\theta})\) from the observed rewards \(r_t\) yields a unique value function decomposition \(V_{\pi_\mathbf{\theta}}(\mathbf{s}) = \frac{r(\pi_\mathbf{\theta})}{1-\gamma} + \tilde{V}_{\pi_{\mathbf{\theta}}}(\mathbf{s}) + e_{\pi_{\mathbf{\theta}}}(\mathbf{s}), \quad \forall \mathbf{s} \in \mathcal{S}\),
where \(r(\pi_\mathbf{\theta})\) is the long-term average reward of \(\pi_\mathbf{\theta}\), defined as \(r(\pi_{\mathbf{\theta}}) \doteq \lim_{T \to \infty} \frac{1}{T} \mathbb{E}_{\pi_{\mathbf{\theta}}} \left[ \sum_{t=0}^{T} r_t \right]\), the state-independent term \(\frac{r(\pi_{\mathbf{\theta}})}{1-\gamma}\) can be ignored during policy improvement, and the differential term \(\tilde{V}_{\pi_{\mathbf{\theta}}}(\mathbf{s}) \doteq \mathbb{E}_{\pi_{\mathbf{\theta}}}\left[\sum_{t=0}^{T}  r_{t} - r(\pi_{\mathbf{\theta}}) \mid \mathbf{s}_0=\mathbf{s}\right]\) provides a stable signal for gradient-based optimization. The error term \(e_{\pi_{\mathbf{\theta}}}(\mathbf{s})\) converges to zero for all states \(\mathbf{s}\) as the \(\gamma \to 1\).
}

\begin{proof}
The stepwise reward $r_t$ can be decomposed as:
\begin{equation}
    r_t = r(\pi_\mathbf{\theta}) + \left(r_t - r(\pi_\mathbf{\theta})\right), \nonumber
\end{equation}
where $r(\pi_\mathbf{\theta})$ is the average reward of policy $\pi_\mathbf{\theta}$. Next, consider an MDP with infinite step and $\gamma<1$. Its discounted value function can be decomposed as:
\begin{align} 
V_{\pi_\mathbf{\theta}}(\mathbf{s}) &=\mathbb{E}_{\pi_{\mathbf{\theta}}}\left[\sum_{t=0}^{\infty}\gamma^t r_t\mid \mathbf{s}_0=\mathbf{s}\right] \nonumber\\
&=\mathbb{E}_{\pi_{\mathbf{\theta}}}\left[\sum_{t=0}^{\infty}\gamma^t\left(r(\pi_\mathbf{\theta}) + \left(r_t - r(\pi_\mathbf{\theta})\right)\right)\mid \mathbf{s}_0=\mathbf{s}\right] \nonumber\\
&= \underbrace{\mathbb{E}_{\pi_{\mathbf{\theta}}}\left[\sum_{t=0}^{\infty}\gamma^t r(\pi_\mathbf{\theta})\mid \mathbf{s}_0=\mathbf{s}\right]}_{\text{constant term}} \nonumber\\
&\quad +\underbrace{\mathbb{E}_{\pi_{\mathbf{\theta}}}\left[\sum_{t=0}^{\infty}\gamma^t\left(r_t - r(\pi_\mathbf{\theta})\right)\mid \mathbf{s}_0=\mathbf{s}\right]}_{\text{differential and error term}}.\nonumber
\end{align}

\noindent The constant term is the infinite discounted sum of average rewards:
$$ \mathbb{E}_{\pi_{\mathbf{\theta}}}\left[\sum_{t=0}^{\infty}\gamma^t r(\pi_\mathbf{\theta})\mid \mathbf{s}_0=\mathbf{s}\right] = r(\pi_\mathbf{\theta})\sum_{t=0}^{\infty}\gamma^t = \frac{r(\pi_\mathbf{\theta})}{1-\gamma}.$$

\noindent We can decompose the differential and error term $\mathbb{E}_{\pi_{\mathbf{\theta}}}\left[\sum_{t=0}^{\infty}\gamma^t\left(R_t - r(\pi_\mathbf{\theta})\right)\mid \mathbf{s}_0=\mathbf{s}\right]$ into a differential term: 
\begin{equation}
    \tilde{V}_{\pi_{\mathbf{\theta}}}(\mathbf{s}) = \mathbb{E}_{\pi_{\mathbf{\theta}}}\left[\sum_{t=0}^{\infty}\left(r_t - r(\pi_\mathbf{\theta})\right)\mid \mathbf{s}_0=\mathbf{s}\right], \nonumber
\end{equation}
\noindent and an error term: 
\begin{equation}
 e_{\pi_{\mathbf{\theta}}}(\mathbf{s}) = \mathbb{E}_{\pi_{\mathbf{\theta}}}\left[\sum_{t=0}^{\infty}(\gamma^t - 1)\left(r_t - r(\pi_\mathbf{\theta})\right)\mid \mathbf{s}_0=\mathbf{s}\right].\nonumber
\end{equation}
Note that as $\gamma$ approaches $1$ the error term $e_{\pi_{\mathbf{\theta}}}(\mathbf{s})$ converges to zero. Combining the constant term, differential term, and error term yields the decomposition:
\begin{equation}
V_{\pi_\mathbf{\theta}}(\mathbf{s}) = \frac{r(\pi_\mathbf{\theta})}{1-\gamma} + \tilde{V}_{\pi_{\mathbf{\theta}}}(\mathbf{s}) + e_{\pi_{\mathbf{\theta}}}(\mathbf{s})\nonumber.
\end{equation}
The constant term $\frac{r(\pi_\mathbf{\theta})}{1-\gamma}$ represents an amplification term for long-term average reward. The differential term $\tilde{V}_{\pi_{\mathbf{\theta}}}(\mathbf{s})$ describes a state's advantage or disadvantage relative to average reward, and the error term $e_{\pi_{\mathbf{\theta}}}(\mathbf{s})$ is the approximation error due to $\gamma<1$.

Now consider the factor-mining MDP with $\gamma=1$ and finite steps. In the decomposition, the constant term $\frac{r(\pi_\mathbf{\theta})}{1-\gamma}$ diverges due to the inappropriate infinite-step summation, and should be redefined as $\mathbb{E}_{\pi_{\mathbf{\theta}}}\left[\sum_{t=0}^{T}  r(\pi_{\mathbf{\theta}})\mid \mathbf{s}_0=\mathbf{s}\right]$, where $T$ denotes the final step. The differential term remains, while the error term vanishes: 
\begin{equation}
    e_{\pi_{\mathbf{\theta}}}(\mathbf{s}) = \mathbb{E}_{\pi_{\mathbf{\theta}}}\left[\sum_{t=0}^{\infty}(\gamma^t - 1)\left(r_t - r(\pi_\mathbf{\theta})\right)\mid \mathbf{s}_0=\mathbf{s}\right]=0. \nonumber
\end{equation}

\noindent Finally the decomposition can be rewritten as:
\begin{equation}
V_{\pi_\mathbf{\theta}}(\mathbf{s}) = \mathbb{E}_{\pi_{\mathbf{\theta}}}\left[\sum_{t=0}^{T}  r(\pi_{\mathbf{\theta}})\mid \mathbf{s}_0=\mathbf{s}\right] + \tilde{V}_{\pi_{\mathbf{\theta}}}(\mathbf{s}) + 0. \nonumber
\end{equation}
\noindent Since the constant term is state-independent, it does not affect relative state values. We can therefore replace the original value estimator with the differential term $\tilde{V}_{\pi_{\mathbf{\theta}}}(\mathbf{s})$. This mean-centered term preserves relative values across states while eliminating the global offset, thereby reducing learning difficulty and improving stability. Specially, the constant tern $\mathbb{E}_{\pi_{\mathbf{\theta}}}\left[\sum_{t=0}^{T}  r(\pi_{\mathbf{\theta}})\mid \mathbf{s}_0=\mathbf{s}\right]$ could become extremely large under a long trajectory. Removing this offset enables neural networks or function approximators to converge efficiently to correct relative values without numerical instability.
\end{proof}

\section{Numerical Results}
\label{Numerical Results}
In this section, a numerical evaluation of TLRS is conducted by benchmarking it against cutting-edge RL algorithms and alternative formulaic alpha-factor mining approaches. The experiments utilize six stock datasets (see Section \ref{Environment Settings}), and assess the performance of different reward shaping algorithms in factor-mining MDP in Section \ref{Comparisons with Other RLs} and \ref{The study of varied gamma}. The effects of hyperparameter settings are investigated in Section \ref{Sensitivity Analysis}. Following this, seven distinct formulaic alpha-factor mining approaches are evaluated in Sections \ref{FacEva}. Finally, an ablation study in Section \ref{Ablation Study} confirms the impact of the two proposed enhancements.

\subsection{Environment Settings}
\label{Environment Settings}
The raw data is sourced from both the Chinese A-shares market and the US stock market. In particular, the constituent stocks of the China Securities Index 300 (CSI300, the index composed of the 300 most liquid and largest A-share stocks listed on the Shanghai and Shenzhen Stock Exchanges), the China Securities Index 500 (CSI500, the index representing 500 A-share stocks with mid-cap market values), the China Securities Index 1000 (CSI1000, an index including 1000 smaller-cap A-share stocks), the S\&P 500 Index (SPX, the Dow Jones Industrial Average, which tracks 30 major US blue-chip companies representing key sectors), the Dow Jones Industrial Average (DJI), and the NASDAQ 100 Index (NDX, the NASDAQ-100, consisting of 100 of the largest non-financial companies listed on the Nasdaq Stock Market) are utilized to model the factor-mining MDP in our experiment. Due to the limited public availability of many macroeconomic, fundamental, and price-volume features, we rely on six key price-volume features for reproducibility when generating our formulaic alphas: opening price (open), closing price (close), highest price (high), lowest price (low), trading volume (volume), and volume-weighted average price (vwap). Our goal is to generate formulaic alphas that demonstrate a high IC with respect to actual 5-day asset returns. The dataset is divided into three segments: a training set covering 01/01/2016 to 01/01/2020, a validation set from 01/01/2020 to 01/01/2021, and a test set from 01/01/2021 to 01/01/2024. Note that all price and volume data have been forward-dividend adjusted based on the adjustment factors as of 01/15/2023.

To assess the performence of TLRS against existing methods, we compare it with tree models, heuristic algorithms, end-to-end deep learning approaches, and reward shaping algorithms in reinforcement learning. Our experiments rely on the open-source implementations available from AlphaGen\cite{yu2023generating}, gplearn\cite{T_Stephens_2015} Stable Baseline 3\cite{raffin2021stable} and Qlib\cite{yang2020qlib}.

\begin{itemize}
    \item Tree Model Algorithms: 
    \begin{itemize}
        \item XGBoost\cite{zhu2022application}: An efficient implementation of gradient boosting decision trees that improves prediction accuracy by combining multiple trees.
        \item LightGBM\cite{li2022research}: Another popular gradient boosting decision tree framework known for its fast performance and low memory usage, suitable for large-scale data analysis.
    \end{itemize}
    \item End-to-End Deep Model Algorithms: 
    \begin{itemize}
        \item MLP\cite{rumelhart1986learning}: A fully connected feedforward neural network adept at capturing complex patterns and nonlinear relationships in data.
    \end{itemize}
    \item Heuristic Algorithms:
    \begin{itemize}
        \item GP\cite{zhang2020autoalpha}: A heuristic search method for complex optimization problems that approximates the optimal solution through iterative generation and evolution of a population of candidate solutions.
    \end{itemize} 
    \item Reward Shaping Algorithms in Reinforcement Learning: 
    \begin{itemize}
        \item No Shaping (NS)\cite{yu2023generating}: A natural solution to the factor-mining MDP, including AlphaGen\cite{yu2023generating}, which utilizes PPO to find interpretable alpha factors. Another advanced algorithm is QFR\cite{11024173}, which proposes a novel improved REINFORCE algorithm.
        \item PBRS\cite{ng1999policy}: It adds a calculated shaping reward $f_{t}=\gamma \Phi\left(\mathbf{s}_{t+1}\right)-\Phi(\mathbf{s}_t)$
 to the original reward of PPO, where the potential function is built on Euclidean distance as $ \Phi(\mathbf{s}_t) = \sqrt{\sum^n_{i=1}\bigl(\mathbf{s}_{t} - \mathbf{s}_{t,i}^e\bigr)^2} $.
        \item DPBA\cite{wiewiora2003principled}: It adds a calculated shaping reward $f_{t}=\gamma \Phi\left(\mathbf{s}_{t+1},a_{t+1}\right)-\Phi(\mathbf{s}_t,a_t)$
 to the original reward of PPO. Because the state transition function in the facor-mining MDP follows
the Dirac distribution, the potential function is built on Euclidean distance as $ \Phi(\mathbf{s}_t,a_t) = \sqrt{\sum^n_{i=1}\bigl(\mathbf{s}_{t+1} - \mathbf{s}_{t+1,i}^e\bigr)^2} $.
 
    \end{itemize}
\end{itemize}

The expert demonstrations (i.e., expert factors) used in the experiment are handcrafted, inspired by Alpha101 \cite{kakushadze2016101}. To account for randomness, each trial is repeated using five distinct random seeds. MLP, XGBoost and LightGBM hyperparameters follow the Qlib benchmark settings, while GP uses the defaults from the gplearn framework. In PPO, the actor and critic share a two-layer LSTM feature extractor (hidden size 128) with a dropout rate of 0.1. Their separate value and policy heads are MLPs with two hidden layers of size 64. The PPO clipping threshold $\epsilon$ is set to 0.2. 
Experiments are run on a single machine equipped with an Intel Xeon Gold 6240R CPU and two NVIDIA RTX A5000 GPUs.


\begin{figure*}[!ht]
\centering
\includegraphics[width=0.95\textwidth]{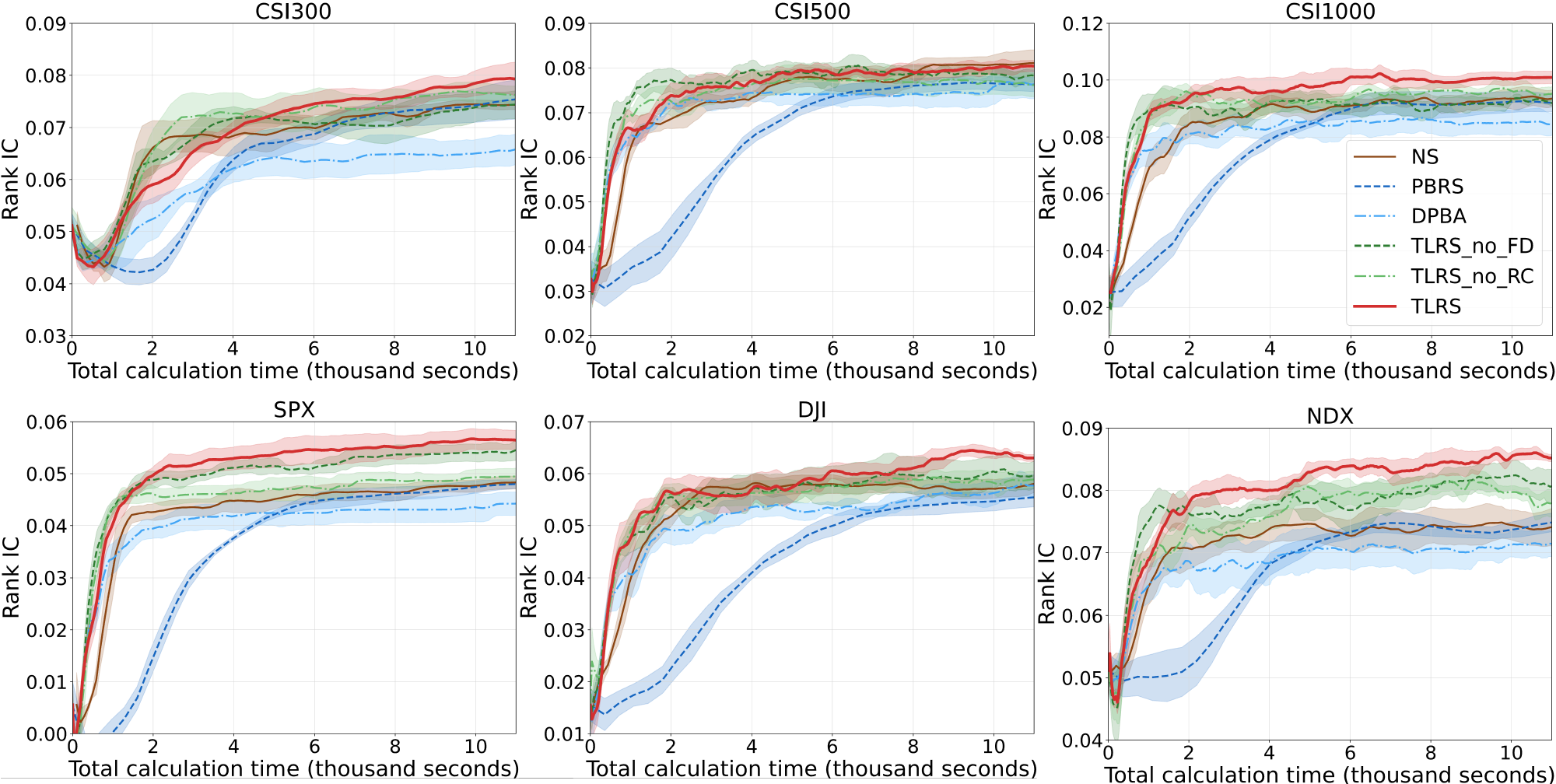}

\caption{Training-phase correlation between mined factor values and the prices of the six index constituents for all investigated reward shaping algorithms and the two variants of TLRS. All the curves are averaged over 5 different random seeds, and half of the standard deviation is shown as a shaded region.}
\label{Figure3}
\end{figure*}

\subsection{Comparisons with Other Reward Shaping Algorithms}
\label{Comparisons with Other RLs}
We present the numerical results of cutting-edge reward-shaping methods in factor mining—namely NS, PBRS, and DPBA—across six market indices (see Fig. \ref{Figure3}). Because TLRS, PBRS, and DPBA incorporate reward shaping but NS does not, we evaluate them using the Rank Information Coefficient (Rank IC) rather than raw reward. Rank IC is defined as $Rank IC\left(\mathbf{z}^{\prime}_t, \mathbf{y}_t\right) = IC\left(r(\mathbf{z}^{\prime}_t), r(\mathbf{y}_t\right))$, where $r(\cdot)$ produces ranks. Higher Rank IC values indicate better factor signal quality. The six indices draw from stocks on the Shanghai, Shenzhen, Nasdaq, and NYSE exchanges, covering a spectrum of company sizes and industry stresses: CSI300 (large caps), CSI500 (mid caps), CSI1000 (small caps), NDX (tech-heavy), DJI (blue-chip leaders), and SPX (broad large-cap U.S. firms). These different constituent stocks may reflect varying levels of market stress in different periods. For example, tech downturns tend to hit NDX harder, while CSI1000 may behave differently in volatile markets. 

As shown in Fig. \ref{Figure3}, apart from the experiment on CSI500 constituents where TLRS and baseline algorithms exhibit similar performance, TLRS achieves faster and more stable convergence, outperforming all other algorithms. It improves Rank IC by 9.29\% over existing potential‐based shaping algorithms. Existing potential‐based shaping algorithms enrich sparse terminal rewards but still suffer from the three issues identified in Section \ref{Challenges for RSfD in Alpha Mining}. They also incur nontrivial overhead: PBRS runs in $\mathcal{O}(N \cdot L \cdot d)$ for Euclidean computations (where $N$ is the number of expert demonstrations, $L$ is the sequence length, and $d$ is the vector dimensionality), and DPBA runs in $\mathcal{O}(N \cdot L \cdot d \cdot P)$ ($P$ denotes the complexity of the policy network’s forward pass). By contrast, TLRS employs exact RPN subsequence matching to eliminate distance‐metric pitfalls, all at $\mathcal{O}(N \cdot L)$ time complexity. This yields faster convergence, and a more efficient factor‐mining process. 

Our dataset spans four years (2016–2020) and covers four exchanges, reflecting a wide spectrum of market conditions and trading stress. QFR maintains superior performance across these scenarios by removing critic-network, employing a subtractive baseline, and applying targeted reward shaping.

\subsection{The Impact of the Discount Factor}
\label{The study of varied gamma}

The impact of the discount factor \(\gamma\) is investigated. The performance of the factor-mining process, specifically validates the theoretical conclusions in Proposition 1 of Section \ref{The Theoretical Analysis}. We conducted experiments using both the standard PPO (NS) algorithm and our proposed TLRS on the CSI300 dataset, varying \(\gamma\) with values of 0.5, 0.8, 0.9, 0.99 and 1. The dataset, by comprising the 300 largest and most liquid A-share stocks that collectively represent over 70\% of the market capitalization, provides a representative foundation for evaluations.

\begin{figure}[!h]
\centering
\includegraphics[width=0.34\textwidth]{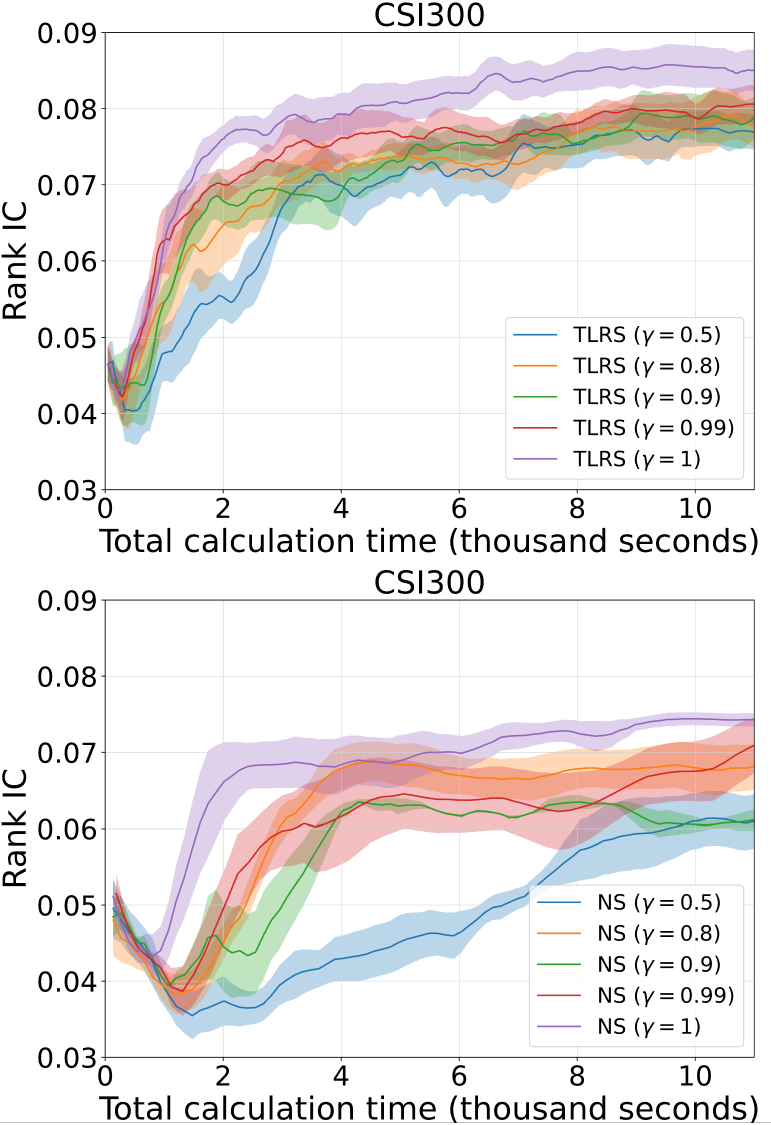}

\caption{Training-phase correlation between mined factor values and the prices of the CSI300 index constituents for the learning time with varied \(\gamma\). All the curves are averaged over 5 different random seeds, and half of the standard deviation is shown as a shaded region.}
\label{Figure4}
\end{figure}

As shown in Figure \ref{Figure4}, both NS and TLRS achieve their best performance when the discount factor \(\gamma\) is set to 1. This empirical evidence strongly supports the theoretical results presented in Proposition 1. When \(\gamma < 1\), the optimization objective (\ref{original objective function}) introduces a bias that encourages the agent to generate shorter factor expressions. This premature termination prevents the discovery of more complex and potentially more predictive factors. By setting \(\gamma = 1\), this bias is eliminated, and the agent is purely motivated to maximize the intrinsic quality of the alpha factor, regardless of the expression's length. This ensures a more thorough exploration of the search space for effective, long-term reward-oriented factors.

\subsection{Sensitivity Analysis}
\label{Sensitivity Analysis}
\begin{figure}[b]
\centering
\includegraphics[width=0.5\textwidth]{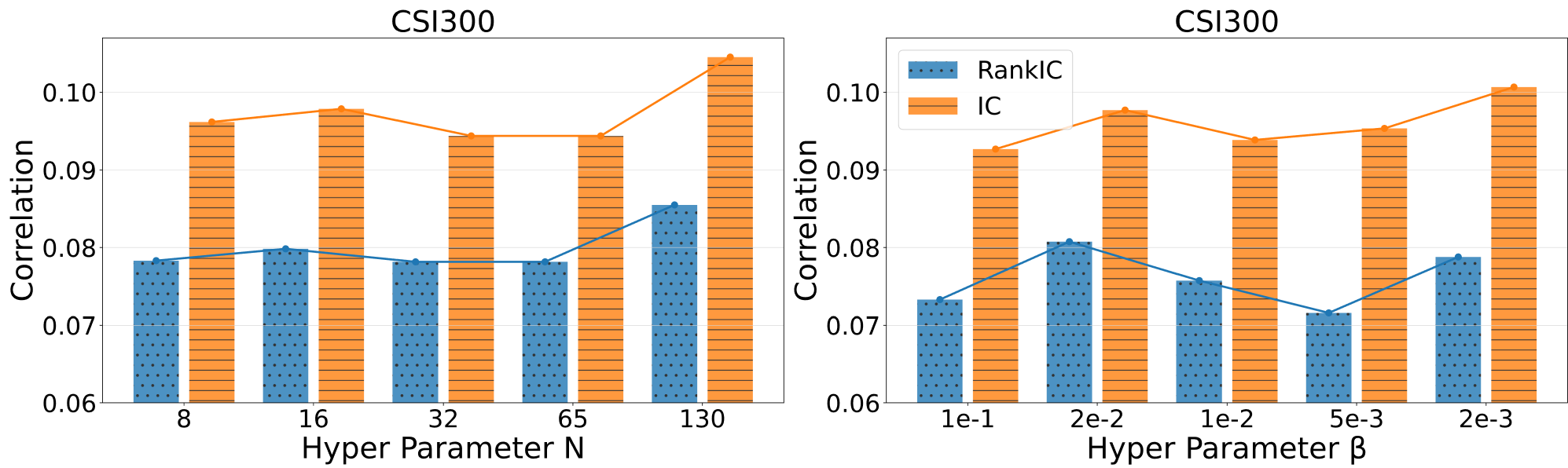}

\caption{Sensitivity analysis of TLRS to number of expert demonstrations \(N\) and the reward centering learning rate \(\beta\).}
\label{Figure5}
\end{figure}

The sensitivity of two key hyperparameters was analyzed: the number of expert demonstrations \(N\) and the reward centering learning rate \(\beta\). As shown in Figure \ref{Figure5}, model performance, measured by Rank IC and IC, consistently improved with more demonstrations. Performance peaked at \(N=130\), which was selected as the optimal value. The impact of \(\beta\) was non-monotonic. The best results were achieved at \(\beta=2e-3\). A smaller \(\beta\) provides a more stable average reward estimate, reducing variance and stabilizing training. We therefore chose \(\beta=2e-3\).

\subsection{Factors Evaluation}
\label{FacEva} 
We evaluates the out-of-sample performance of the alpha factors generated by TLRS against those from several baseline algorithms. The evaluation is conducted on the CSI300 and CSI500 indices, with performance measured by IC and RankIC. The results, including the mean and standard deviation over five runs, are summarized in Table \ref{Performance of Mined Factors on CSI300 and CSI500}.

The results demonstrate the competitive performance of TLRS. On both the CSI300 and CSI500 indices, TLRS achieves IC and RankIC scores that are on par with or exceed those of the top-performing baseline methods, but it does not demonstrate a significant statistical superiority over other state-of-the-art methods like QFR. While TLRS shows faster convergence during training (as seen in Figure \ref{Figure3}), its final predictive power is comparable to the best baselines. One possible explanation for this is that the performance of all evaluated algorithms, including TLRS, might be approaching a performance ceiling imposed by the limited feature set. With only six basic price-volume features as inputs, there may be a natural limit to the predictive signal that any formulaic factor can extract. The fact that several distinct algorithms converge to a similar performance level suggests that the bottleneck may lie more in the informational content of the raw data than in the factor generation algorithm itself.

However, when comparing TLRS with AlphaGen, which is also based PPO, the innovative advantages of TLRS become particularly prominent. The results demonstrate the clear superiority of TLRS. This strong performance stems from two core innovations. First, the trajectory-level reward shaping effectively guides the agent's exploration by leveraging expert demonstrations. Unlike those computationally intensive, distance-based shaping methods, TLRS's subsequence matching provides dense, intermediate signals, steering the policy toward structures found in successful alpha factors. Second, the reward centering mechanism reduces the high variance inherent in the learning process, leading to more stable and efficient convergence. This combination of intelligent guidance and enhanced training stability allows TLRS to navigate the vast search space more effectively, ultimately discovering more robust and predictive alpha factors than other heuristic or standard RL algorithms.
\begin{table}[t]
\centering
\caption{Testing-Phase Correlation Between the Prices of CSI300/CSI500 Constituents and Factor Values Mined by All Investigated Factor Mining Algorithms}
\label{Performance of Mined Factors on CSI300 and CSI500}
\begin{tabular*}{0.48\textwidth}{@{\extracolsep{\fill}}c|cc|cc}
\toprule
\multirow{2}{*}{}   & \multicolumn{2}{c|}{CSI300} & \multicolumn{2}{c}{CSI500}  \\ \cline{2-5} 
                          & IC           & RankIC        & IC  & RankIC      \\ \hline
\multirow{2}{*}{MLP}      & 0.0123                 & 0.0178                 & 0.0158       & 0.0211               \\
                          & (0.0006)               & (0.0017)      & (0.0014)     & (0.0007)             \\ \hdashline
\multirow{2}{*}{XGBoost}  & 0.0192                 & 0.0241                 & 0.0173       & 0.0217               \\
                          & (0.0021)               & (0.0027)               & (0.0017)     & (0.0022)             \\ \hdashline
\multirow{2}{*}{LightGBM} & 0.0158                 & 0.0235                 & 0.0112       & 0.0212               \\
                          & (0.0012)               & (0.0030)               & (0.0012)     & (0.0020)             \\ \hdashline
\multirow{2}{*}{GP}       & 0.0445                 & \textbf{0.0673}        & 0.0557  & 0.0665               \\
                          & (0.0044)               & \textbf{(0.0058)}      & (0.0117) & (0.0154)             \\ \hdashline
\multirow{2}{*}{AlphaGen} 
                          & 0.0500        & 0.0540                & 0.0544       & \textbf{0.0722}               \\
                          & (0.0021)      & (0.0035)               & (0.0011)     & \textbf{(0.0017)}             \\ \hdashline
\multirow{2}{*}{QFR}    &\textbf{0.0588}  & 0.0602        & \textbf{0.0708}     &  0.0674     \\
                          &  \textbf{(0.0022)}      &  (0.0014)      &  \textbf{(0.0063)}    & (0.0033)  \\ \hline
\multirow{2}{*}{TLRS}    &\textbf{0.0571}  & \textbf{0.0582}        & \textbf{0.0717}     &  \textbf{0.0730}     \\
                          &  \textbf{(0.0096)}      &  \textbf{(0.0128)}      &  \textbf{(0.0143) }   & \textbf{(0.0097)}  \\ \bottomrule
\end{tabular*}
\end{table}

\subsection{Ablation Study}
\label{Ablation Study}
To isolate the impact of the two improvements of TLRS, we designed two variants: TLRS without reward shaping (TLRS\_no\_RS) and TLRS without reward centering (TLRS\_no\_RC). As shown in Figure \ref{Figure3}, the complete TLRS algorithm consistently outperforms both variants across all datasets, confirming that both components are crucial. The performance degradation in TLRS\_no\_RS highlights that reward shaping is essential for guiding exploration effectively with sparse rewards. Similarly, the lower performance of TLRS\_no\_RC demonstrate that reward centering is vital for stabilizing the training process. In summary, the results confirm that trajectory-level reward shaping and reward centering are complementary and indispensable for the superior performance of TLRS.


\section{Conclusion}
\label{Conclusion}
In this paper, we have proposed Trajectory-level Reward Shaping (TLRS), a novel and effective algorithm for mining formulaic alpha factors. TLRS addresses the unique challenges of the factor-mining MDP, such as sparse rewards and semantic ambiguity, by introducing a novel reward shaping mechanism based on exact subsequence matching with expert demonstrations. Furthermore, it incorporates a reward centering technique to mitigate high reward variance and enhance training stability. Our extensive experiments on diverse, real-world stock market datasets demonstrate that TLRS achieves competitive performance against state-of-the-art RL algorithms and traditional factor mining methods, generating alpha factors with strong predictive power. We conclude that TLRS is a powerful and promising approach for discovering high-quality, interpretable alpha factors. Future work could involve integrating the large language models to enhance TLRS's capability in capturing cross-stock correlations.

\bibliographystyle{IEEEtran}
\bibliography{mybib.bib}

\newpage

\end{document}